\documentclass{article}
\usepackage{spconf,amsmath,graphicx}
\usepackage{booktabs}
\usepackage[colorlinks,
linkcolor=black,
citecolor=black,
anchorcolor=black,
urlcolor=black]{hyperref}
\title{Face Mask Assistant: Detection of Face Mask Service Stage Based on Mobile Phone}
\name{Yuzhen Chen$^{1}$, Menghan Hu$^{1}$, Chunjun Hua$^{1}$, Guangtao Zhai$^{2}$, Jian Zhang$^{1}$,Qingli Li$^{1}$, Simon X. Yang$^{3}$}
\address{$^1$Shanghai Key Labora. of Multidim. Infor. Proce., East China Normal University, China\\
$^2$Key Laboratory of Artiﬁcial Intelligence, Ministry of Education, China\\
$^3$Advanced Robotics and Intelligent Systems Laboratory, School of Engineering, University of Guelph\\
\thanks{This work is sponsored by the National Natural Science Foundation of China (No. 61901172, No. 61831015, No. U1908210), the Shanghai Sailing Program (No.19YF1414100), the “Chenguang Program” supported by Shanghai Education Development Foundation and Shanghai Municipal Education Commission (No. 19CG27), the Science and Technology Commission of Shanghai Municipality (No. 19511120100, No. 18DZ2270700, No. 14DZ2260800), the foundation of Key Laboratory of Artificial Intelligence, Ministry of Education (No. AI2019002), and the Fundamental Research Funds for the Central Universities.}\\
\thanks{Corresponding author: Menghan Hu (mhhu@ce.ecnu.edu.cn)}}

\begin{document}
%
\maketitle

\begin{abstract}
Coronavirus Disease 2019 (COVID-19) has spread all over the world since it broke out massively in December 2019, which has caused a large loss to the whole world. Both the confirmed cases and death cases have reached a relatively frightening number. Syndrome coronaviruses 2 (SARS-CoV-2), the cause of COVID-19, can be transmitted by small respiratory droplets. To curb its spread at the source, wearing masks is a convenient and effective measure. In most cases, people use face masks in a high-frequent but short-time way. Aimed at solving the problem that we don't know which service stage of the mask belongs to, we propose a detection system based on the mobile phone. We first extract four features from the GLCMs of the face mask's micro-photos. Next, a three-result detection system is accomplished by using KNN algorithm. The results of validation experiments show that our system can reach a precision of 82.87\%$\pm$8.50\% on the testing dataset. In future work, we plan to expand the detection objects to more mask types. This work demonstrates that the proposed mobile microscope system can be used as an assistant for face mask being used, which may play a positive role in fighting against COVID-19.

\end{abstract}
\section{Introduction}
\label{sec:intro}
COVID-19 has wreaked havoc around the world and the spread of COVID-19 has not been curbed totally by June, 2020. According to the situation report of the WHO, there have been 9,653,048 confirmed cases and 491,128 deaths by 27 June 2020\cite{world2020coronavirus}. Recently, many researchers have developed a variety of detection methods of COVID-19, such as reverse transcription-polymerase chain reaction (RT-PCR) technology\cite{lan2020positive, hussein2020point}, X-ray\cite{wang2020covid} and chest CT imaging\cite{bernheim2020chest}. In addition to the need for breakthroughs in testing techniques, SARS-CoV-2 must be controlled at source to prevent a sustaining surge in COVID-19 confirmed and death cases. Small respiratory droplets can transmit SARS-CoV-2, which is known to be transmissible from presymptomatic and asymptomatic individuals\cite{howard2020face}. One thing to reduce virus's spreading is wearing masks in public\cite{howard2020face}. Advice on using face masks proposed by the WHO\cite{10665-332293}, analysis\cite{howard2020face} and researches\cite{greenhalgh2020face} have supported that wearing masks can prevent the spread of potential viruses during pandemic. An evidence research even thinks that wearing masks in public is a symbol of social solidarity among the response to the worldwide pandemic \cite{cheng2020wearing}. The WHO recommends that the surgical face masks used by health workers should be discarded once used and during severe shortages the extended use of faces masks can be considered \cite{world2020rational}. Because face masks only work when being worn within their periods of validity. On the other hand, more often in daily life, the majority of people wear surgical face masks in short-time but high-frequent way and in low-risk or middle-risk environment. The face masks may be used twice or three times, even four or five times if in relatively safe environment and maintained properly, which is reasonable and common especially under the condition where face masks are in shortage.

In this case, there is no doubt that the service life of masks depends on the use methods, use intensity, environmental condition and the material of masks, etc. In daily life, it often happens that common people forget the service stage of the current mask after having used it in the high-frequent but short-time way. Meanwhile, the service life of face masks decreases differently after being used in different conditions. As a result, the masks' remaining time of effective protection varies. With those faces masks whose use time are unknown, continuing to use them may lead to poor protection and other unexpected incidents; discarding them directly is an inappropriate behavior. It is a waste of face masks, especially in the case that masks are in shortage during the epidemic. It is necessary to detect the service stage of our face masks, but there are few researches on the service stage of face masks. Therefore, we aim at studying what service stage the face mask is in, which may play a positive role in protecting the uninfected people and reducing the spread of virus.

According to Javid et al., wearing a mask in public may become our unified action in the fight against COVID-19\cite{javid2020covid}. Due to the short service life and worldwide use of masks, it is not practical for masks to adopt special machines to centralized detection. Hence, it is urgently needed to develop an easy-to-operate, portable testing device for detecting anytime and anywhere. Therefore, in this research, we propose a portable mask service stage detection system based on mobile phone. Its work procedures are as follows: 1) users first take a micro-photo of the mask being used with a mobile microscope; 2) then the micro-graph of the mask is uploaded in the WeChat Applet; 3) the result of back-end detection eventually can be obtained through WeChat Applet. This detection device is simple, easy to operate and effective.

In the common fight, the ordinary people often wear surgical masks while N95 masks are only recommended for healthcare workers or professionals at high risk of coming into contact with patients\cite{long12381effectiveness}. Radonovic et al. analyzed the data which came from 7 health care delivery systems and 4 seasons of peak viral respiratory illness, and found that in this trial, wearing N95 respirators and wearing medical masks resulted in no significant difference in the incidence of laboratory-confirmed influenza\cite{radonovich2019n95}. Additionally, the research suggests that N95 respirators should not be recommended for the general public, namely, those not in close contact with influenza or suspected patients\cite{long12381effectiveness}. Using N95 respirators may lead to discomfort, like headaches\cite{cowling2010face}. Even a previous study \cite{chen2017herd} has reported that there is an inverse relationship between the level of wearing N95 respirators and the risk of clinical respiratory illness. Therefore, we choose surgical mask as the research object, which is the most widely used. It makes experimental data more scientific and ultimately our detection system can benefit more people to the greatest extent. In this study, after obtaining the micro-graphs of masks at different wearing periods, the texture features of the photos are analyzed.

Methods of texture analysis include fractal analysis, Fourier transformation and the gray level co-occurrence matrix (GLCM)\cite{kanai2020discriminant}. Haralick et al. reported that the GLCM was a method to quantify the spatial relationship between adjacent pixels in an image\cite{haralick1973textural}. GLCM is widely used in disease detection \cite{gibbs2003textural,rathore2014ensemble}, skin texture analysis\cite{ou2014vivo}, and defect detection\cite{shabir2019tyre}, etc. Given the above, GLCM is feasible and efficient in image texture feature analysis. Therefore, we choose GLCM to analyze the texture features of masks.

Due to K Nearest Neighbor (KNN) method's simple implementation and distinguished performance in classification tasks\cite{zhang2017learning}, it is widely employed in text categorization\cite{du2019parallel,li2019improved}, medical domain\cite{hamed2020accurate}, eating patterns exploration\cite{newby2004empirically}, leaf disease identification\cite{krithika2017individual}, indoor localization\cite{hoang2018soft}, etc. Hamed et al. compared the performance of KNN variant (KNNV) algorithm (which derived from KNN) and other three algorithms in the classification of COVID-19 patients, and found that KNNV algorithm could classify COVID-19 patients more efficiently and accurately\cite{hamed2020accurate}. On the basis of the texture features extracted from the micro-photos taken in different period time, this paper employs KNN algorithm to establish the classfiers for detecting the use time of mask. 

In this paper, we propose a portable face mask service stage detection system, which is easy to operate and can work anytime and anywhere. It is based on the texture features of face masks and ground-truth of mask service stage data. To further quantify the differences of masks in different periods, the method of texture analysis is introduced. After extracting texture features from the micro-photos of face masks in different use periods, KNN algorithm is performed to detect which service period the face mask belongs to. The testing data are afterward used to validate the effectiveness of the proposed system.

The main contributions of this paper are threefold. First, we combine the texture features of face masks' micro-photos and their corresponding service stages based on the method of texture analysis. Based on GLCM, the texture features are successfully extracted from micro-photos obtained by high magnification lens attached to the mobile phone. Subsequently, we propose a detection method to judge the service stage of face masks with a KNN algorithm. Finally, based on the two contributions mentioned above, we have implemented a detection Wechat Applet for face masks using the data collected from three persons' wearing face masks in daily life, which may contribute to protecting the uninfected people and reducing the spread of virus.

\section{Methods}
\label{sec:method}

A brief introduction to the proposed face mask service stage detection method is shown below. We initially use the portable and delicate mobile microscope device to get the micro-photos of face masks. After obtaining the picture, the first step is to extract texture features data from it. During the extraction process, we use GLCM to capture the contrast, energy, correlation, and homogeneity of the object face mask. Subsequently, KNN algorithm is employed to propose a classification model with the input texture features and the corresponding ground-truth use time. Eventually, we package the system into a library function in Matlab and embed it to the Wechat Applet we develop. This has been proved effective by similar previous researches\cite{jiang2020detection,imran2020ai4covid}.

\subsection{Overview of Data Acquisition}
\begin{figure}[ht]
	\centering
	\includegraphics[width=0.45\textwidth]{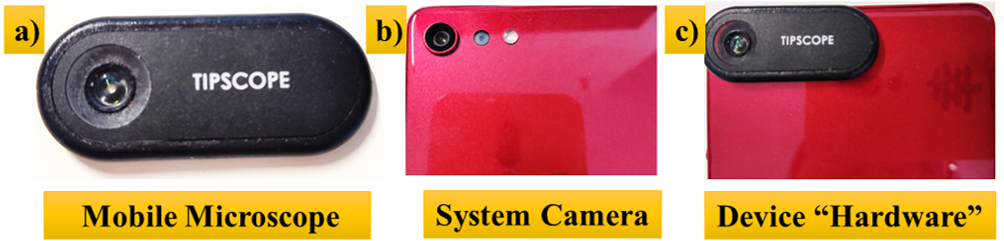}
	\caption{The overview of the portable detection device's ``Hardware''. The ``Hardware'' is composed of a mobile microscope, a phone with available system camera, etc. Specifically, a) and b) represent the mobile microscope and system camera, respectively; and c) shows the final removable device mainly assembled by a) and b).}\label{F01}
\end{figure}

\begin{figure}[ht]
	\centering
	\includegraphics[width=0.5\textwidth]{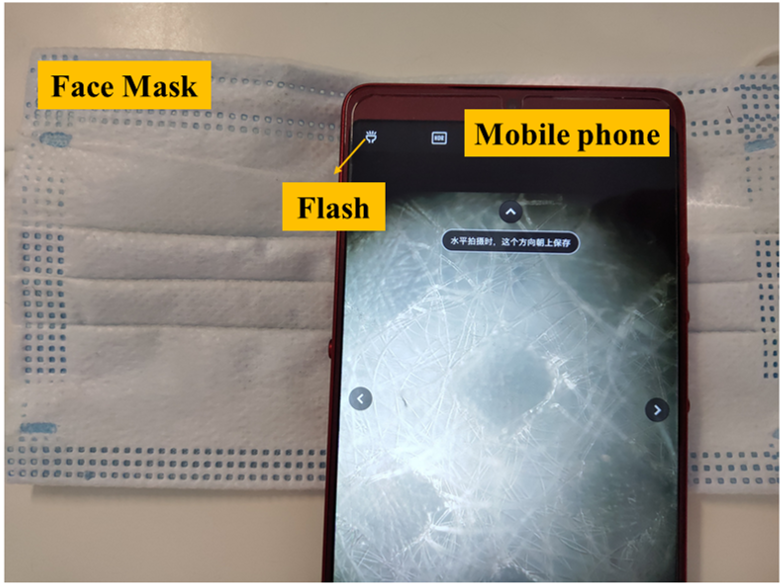}
	\caption{Scenario of the portable detection device's working. While the device is on work, the flash must meet the condition that the flash is always ``on'' to provide enough light.}\label{F02}
\end{figure}

\begin{figure}[ht]
	\centering
	\includegraphics[width=0.5\textwidth]{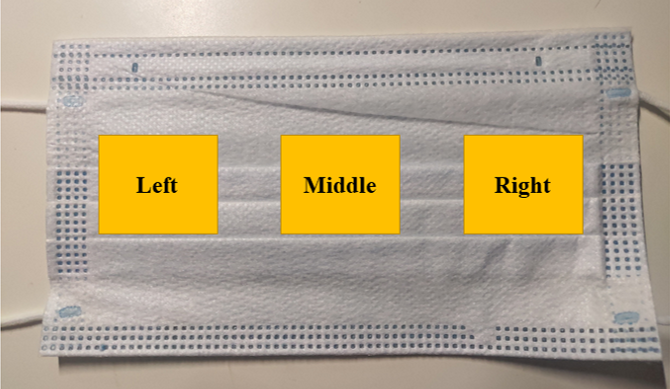}
	\caption{The three inside photographed locations of the face mask: left location (about one-third of the face mask and close to the left), middle location (about two-thirds of face mask), right (about one-third of the face mask and close to the right).}\label{F13}
\end{figure}

Fig. \ref{F01} displays the hardware configuration of the detection device: a mobile microphone (TIPSCOPE, Convergence(Wuhan) Technology Ltd., China), a phone with available system camera, etc. The detachable mobile microphone, which is the key to the hardware, is particularly chosen to display the details of face masks viz. ash particle, droplets or other debris more explicitly. It should be noted that the maximum magnification can reach 400x if the system camera and mobile microscope work together. In this research, we use system camera without magnification to get more scope of face masks. 
Fig. \ref{F02} shows how to collect the experimental data. To ensure the validity of the collected data, we photograph three inside locations of the face mask each time: left location (about one-third of the face mask and close to the left), middle location (about two-thirds of face mask), right (about one-third of the face mask and close to the right), as shown in Fig. \ref{F13}. In order to contain more usage scenarios, the face masks are worn in different conditions. Meanwhile, the flash of the phone's camera should be always ``on'' to provide enough light. We collect the images of a face mask used from day zero (new) to day five.
Based on the conditions mentioned above, three images are immediately photographed after the face mask has been used for an arbitrary hour of the daytime.

\subsection{Gray Level Co-Occurrence Matrix (GLCM)}
Texture can be used to characterize the tonal or gray-level variations in an image\cite{hall2017glcm}. The gray level co-occurrence matrix (GLCM) is chosen to disclose the texture features hiding in the face masks. GLCM is a second-order statistical method which calculates the frequency of pixel pairs with the same gray-level in an image and uses additional knowledge obtained from spatial pixel relations\cite{wu2015active}. The co-occurrence matrix uses edge information to embed the distribution of gray-scale transformation\cite{xing2019multilevel}. Owing to the fact that much of the information required is embedded in GLCM, it emerges as a simple but effective technique.

14 measures of textural features are proposed by Haralick et al., and some of these measures are related to specific textural characteristics of the image\cite{haralick1973textural}. In our study, we choose contrast, correlation, energy and homogeneity as measures. Specifically, contrast refers to the drastic change in gray level between adjacent pixels. High contrast images own high spatial frequencies. Correlation represents the linear correlation in an image. The higher the correlation value, the more linear the gray-scale relationship between adjacent pixel pairs. Energy stands for texture uniformity or pixel pair repetitions. High energy is produced when the distribution of gray level values is constant or periodic. Homogeneity is sensitive to the existence of near diagonal elements in a GLCM, indicating the similarity of adjacent pixels in gray scale.

With the increase of the use times, the wear of the mask causes a change in micro-structure of the mask's some parts, which leads to the difference between the worn-out parts and the unworn parts. Thus, the images in different the mask's service stage
may perform variously in correlation and energy. Meanwhile, due to humans' respiration, cough and other respiratory related activities, more and more small particles impurities and droplets will adhere to the inside surface to the face mask. They probably result in the difference in the four measures mentioned above, among the images of different service stages. 

Every element in GLCM contains second-order statistics, probability values for changes between gray levels $i$ and $j$ for a particular displacement $d$ and angle $\theta$, labeled as $p(i,j)$ (normalized)\cite{lofstedt2019gray}. Let $M$ be the number of gray levels in the image, thus, the size of GLCM is $M\times M$. In this research, we set $M$ as $8$. Subsequently, the measures are calculated as

\begin{equation}\label{E01}
Contrast=\sum_{i=0}^{M-1} \sum_{j=0}^{M-1} (i-j)^2 \cdot p^2(i, j)
\end{equation}

\begin{equation}\label{E02}
Homogeneity=\sum_{i=0}^{M-1} \sum_{j=0}^{M-1} \frac{1}{1+(i-j)^2} \cdot p(i, j)
\end{equation}

\begin{equation}\label{E03}
Energy=\sqrt{\sum_{i=0}^{M-1} \sum_{j=0}^{M-1} p^2(i, j)}
\end{equation}

\begin{equation}\label{E04}
Correlation=\sum_{i=0}^{M-1} \sum_{j=0}^{M-1} (i-\mu)\cdot (j-\mu)\cdot p(i, j)/\sigma^2
\end{equation}

where $\mu$ is the mean of GLCM while $\sigma^2$ stands for the variance of GLCM.

\subsection{K Nearest Neighbor (KNN) Algorithm}
Considering that this task is relatively simple, so the model can be constructed with a relatively simple pattern recognition method. At the same time, in view of the deployment to the mobile phone terminal, the established model should be easy to deploy to the mobile phone terminal and convenient to maintain later. Compared with other algorithms, due to its simple implementation and significant classification performance, KNN is a very popular method in statistics and ranks top ten data mining algorithms{\cite{zhang2018efficient, 7401007, wu2007top,zhang2012nearest}.
	
KNN algorithm is different from model-based methods which first use training samples to build a model, and then predict the test samples through the learned model\cite{yu2014a, zhu2015targeting,shao2014feature}. No training phase is required for the model-free KNN method. Instead, it conducts classification tasks obeying the following procedures: first, training samples are attached labels; next, the distance is calculated between the test sample and the training samples in each label; subsequently, after comparing the distances and obtaining the test's nearest neighbors, the serial number is obtained; finally, classification is generated, as shown in Fig. \ref{F04}. The distance is computed as:
\begin{equation}\label{E05}
	Distance=\sqrt{\sum_{j=1}^{n} (test\ sample-training\ sample_{j})^2} 
\end{equation}
where the $n$ stands for the number of labels.
	
\begin{figure*}[ht]
		\centering
		\includegraphics[width=0.9\textwidth]{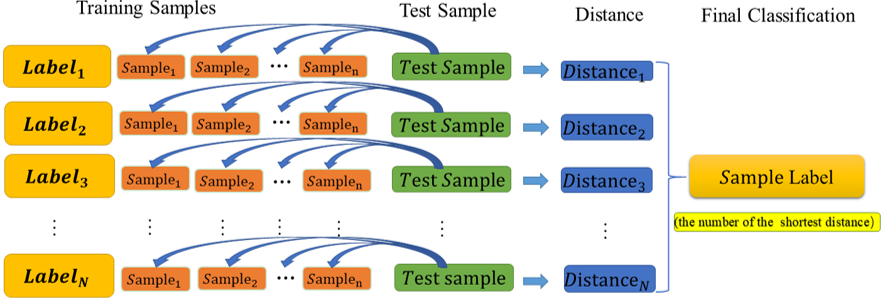}
		\caption{The flowchart of KNN algorithm. First, labels are attached to the training  samples. For every test sample, the $distance_{i}$ is calculated between the test sample and the training samples in each label, which follows the equation \ref{E05}. By comparing the distances, the test sample's nearest neighbors viz.the minimum distance are obtained, and serial number $i_{min}$ is then produced. Eventually, the final classification is set as label $i_{min}$.}\label{F04}
\end{figure*}
	
The main task in our research is to train the optimal $k$ value after carrying out KNN classification successfully according to the efficiency of the classification performance.
	
\subsection{Portable device based on mobile phone for detecting use time of mask}
After optimizing the system, the system is embedded to the Wechat Applet we have developed. To enrich our experimental data, we enlarge the data collection time from the first day to the fifth day (about an hour used time each day). It is generally recommended to replace the surgical mask every four hours. If the face mask is used for a long time, the large particles will be blocked on the mask surface or the ultrafine particles will be blocked in the pores of the mask filter material, resulting in the decrease of filtration efficiency and the increase of respiratory resistance\cite{shiyong}. According to the detection time, the detection results are within three categories viz. type I: normal use (day 0 to 1 viz. the face masks can be used securely), type II: early warning (day 2 to 3 viz. the face masks can be used safely, but it is close to the end of its service life) and type III: not recommended (day 4 to 5 viz. the face mask can be used in shortage, but it is better to change a new one). We choose the `day' as unit based on the fact that we use face masks in high-frequent but short-time way and ordinary people hardly use face masks for 4 hours continuously. Additionally, on basis of using the same face mask several times, we classify the face masks belonging to day 4 and day 5 the type `not recommended' to counterbalance the increase in use times.  
\begin{figure*}[ht]
	\centering
	\includegraphics[width=1\textwidth]{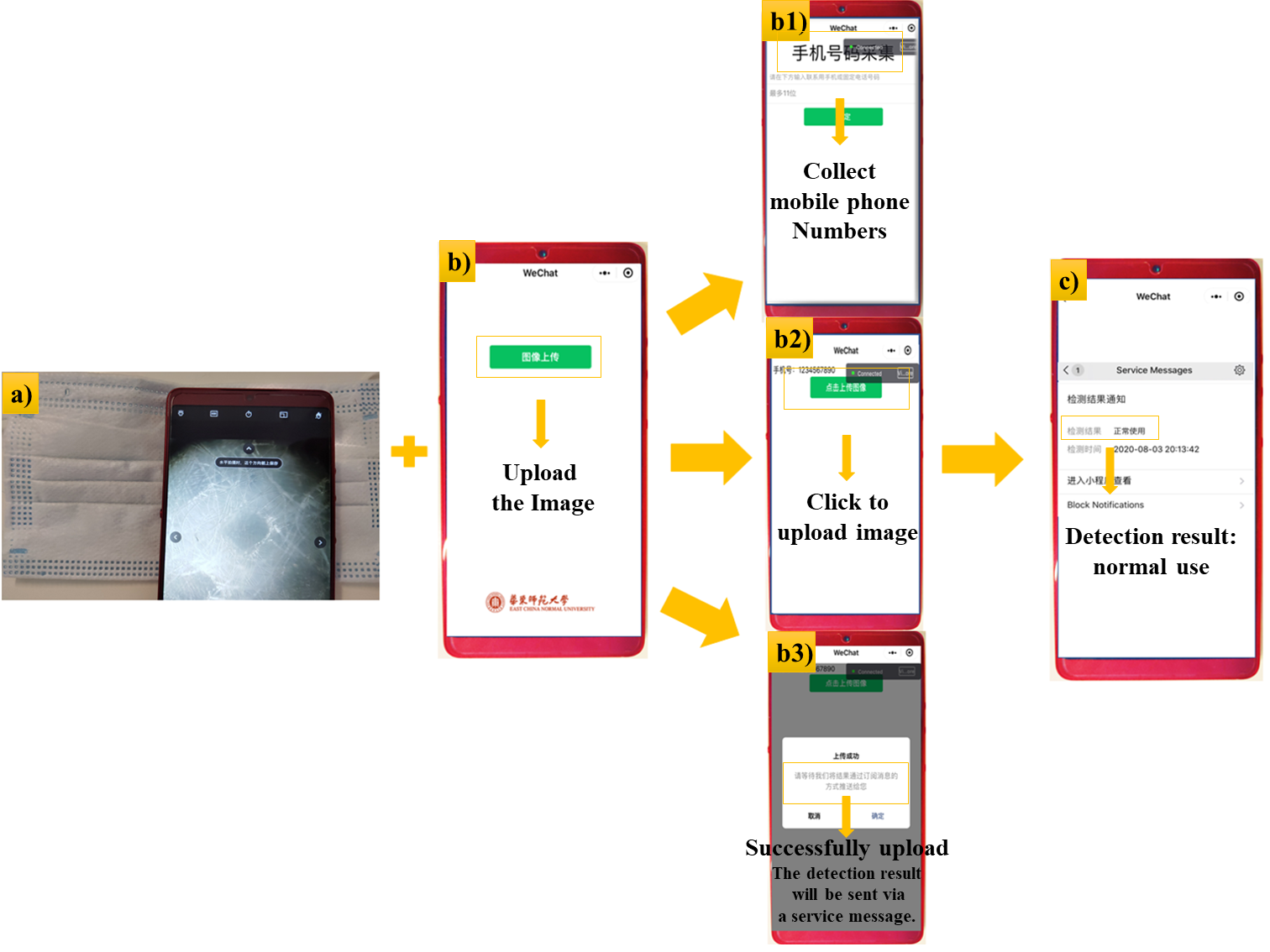}
	\caption{The workflow chart of the face mask detection system. a) stands for collecting the micro-image of the face mask being used. b) with its branches b1), b2) and b3) means registering an account used to receive the detection result and the following uploading task. c) is the result sent by the Wechat Applet from the back-end. (The face mask used in the sample is in its service life of type I: normal use.)}\label{F03}
\end{figure*}
Fig. \ref{F03} reveals the workflow of the detection. After we have obtained the micro-photo of the face mask with mobile microscope and registered an account in the Wechat Applet, the micro-image is uploaded to the back-end of the system. Accordingly, the detection result will be sent in the message from the back-end.

\subsection{Evaluation Metrics}
To verify the efficiency of the detection system, macro-measures viz. macro precision, macro recall, and macro $F_{1}$ are considered. 

1) Confusion matrix: we assume that ``Positive'' means the positive samples and ``Negative'' means the negative samples. Meanwhile, ``True'' represents that the prediction is right while ``False'' represents that the prediction is wrong. As a result, ``TP'' and ``TN'' mean that the positive sample is classified as ``Positive'' and the negative sample  is labeled as ``Negative'', respectively. ``FP'' and ``FN'' represent that the negative sample is labeled as ``Positive'' and the positive sample is classified as  ``False''. The four indicators make up the confusion matrix.

2) Precision: precision is only used to evaluated the classification ability of the positive samples within the range from 0 to 1. It is obvious that the larger precision is, the more effective the system is. It is computed by:
\begin{equation}\label{E07}
Precision=\frac{TP}{TP+FP}
\end{equation}

3) Recall: it is a ratio from 0 to 1. Obviously, the more it is close to 1, the better the system is. The calculation equation is:
\begin{equation}\label{E08}
Recall=\frac{TP}{TP+FN}
\end{equation}

4) $F_{1}$: it is a harmonic mean of recall and precision. In this study, we consider the weight of recall and precision the same, which means attaching the weight of 0.5 to either of them. It is calculated by:
\begin{equation}\label{E09}
F_{1}=\frac{2*Precision*Recall}{Precision+ Recall}
\end{equation}

7) Macro-measures: metrics are calculated for each label, and then the weighted means are produced (normally the weight is the same except that the number of the samples in each label varies greatly). 
\begin{equation}\label{E13}
Macro\_Precision=\frac{1}{n}\sum_{i=1}^{n}Precision_{i}
\end{equation}

\begin{equation}\label{E14}
Macro\_Recall=\frac{1}{n}\sum_{i=1}^{n}Recall_{i}
\end{equation}

\begin{equation}\label{E15}
Macro\_F_1=\frac{1}{n}\sum_{i=1}^{n}Macro\_{F_1}_{i}
\end{equation}
where $n$ means the total number of labels, and the $Macro\_{F_1}_{i}$ is $F_{1}$ of every label calculated by Equation \ref{E09}.
\section{Experimental Results and Analysis}

In this study, we have collected 87 micro-images (excluding the invalid data) of surgical masks via the mobile microscope. The life span of face masks have little relationship with different people but much with the condition of use environment. Meanwhile, people's doing when wearing face masks also influences the life span of face masks. As a result, the data cover the conditions: speaking frequently, talking with others, running, shopping in supermarkets, shopping in vegetable markets, riding and wandering, etc.  

Every time three micro-photos (the left, the middle and the right) are photographed, they are initially averaged to eliminate existing disturbance, and then one final training sample is produced. As a result, after extracting texture features, 29 sets of training samples are obtained. The original 87 micro-images are used to carry out validation experiments. The evaluation indicators mentioned above are elaborated as the follows.

\subsection{Experimental Results}
\begin{figure}[ht]
	\centering
	\includegraphics[width=0.45\textwidth]{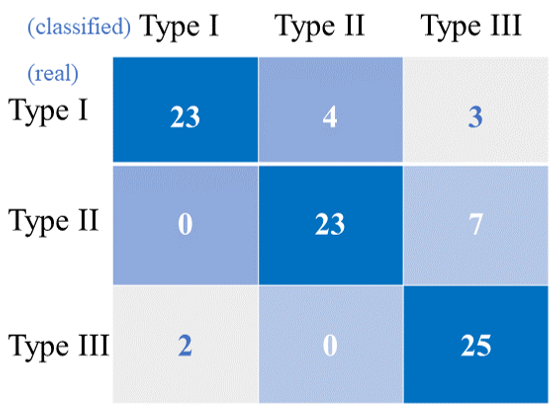}
	\caption{The confusion matrix of the classified results. Each row is the number of real labels and each column is the number of classified labels. Type I, Type II and Type III represent `normal use', `early warning' and `not recommended', respectively.}\label{F05}
\end{figure}
To figure out the detailed information about the classification of the use time, we plot the confusion matrix of the three types as demonstrated in Fig. \ref{F05}. As can be seen from the results, 71 samples among 87 samples are predicted correctly, which suggests the model performs relatively well on the whole. 

\begin{table}[htbp]
	\centering
	\caption{The Macro Measures of Type I, II and III.}
	\begin{tabular}{rrcccc}
		\toprule
		&       & \multicolumn{2}{c}{Macro Precision} & \multicolumn{2}{c}{82.87\%$\pm$8.50\%} \\
		\multicolumn{2}{c}{Macro measures} & \multicolumn{2}{c}{Macro Recall} & \multicolumn{2}{c}{81.98\%$\pm$7.50\%} \\
		&       & \multicolumn{2}{c}{Macro $F_1$} & \multicolumn{2}{c}{81.66\%$\pm$1.40\%} \\
		\bottomrule
	\end{tabular}%
	\label{tab1}%
\end{table}%
According to the equations of macro-measures, we can easily see that macro-measures are convincing and scientific. The macro-precision of the system reaches 82.87\%$\pm$8.50\%, which proves the efficiency of the detection system. 
\subsection{Influence of K Factor on the Efficiency of System}
\begin{figure}[ht]
	\centering
	\includegraphics[width=0.45\textwidth]{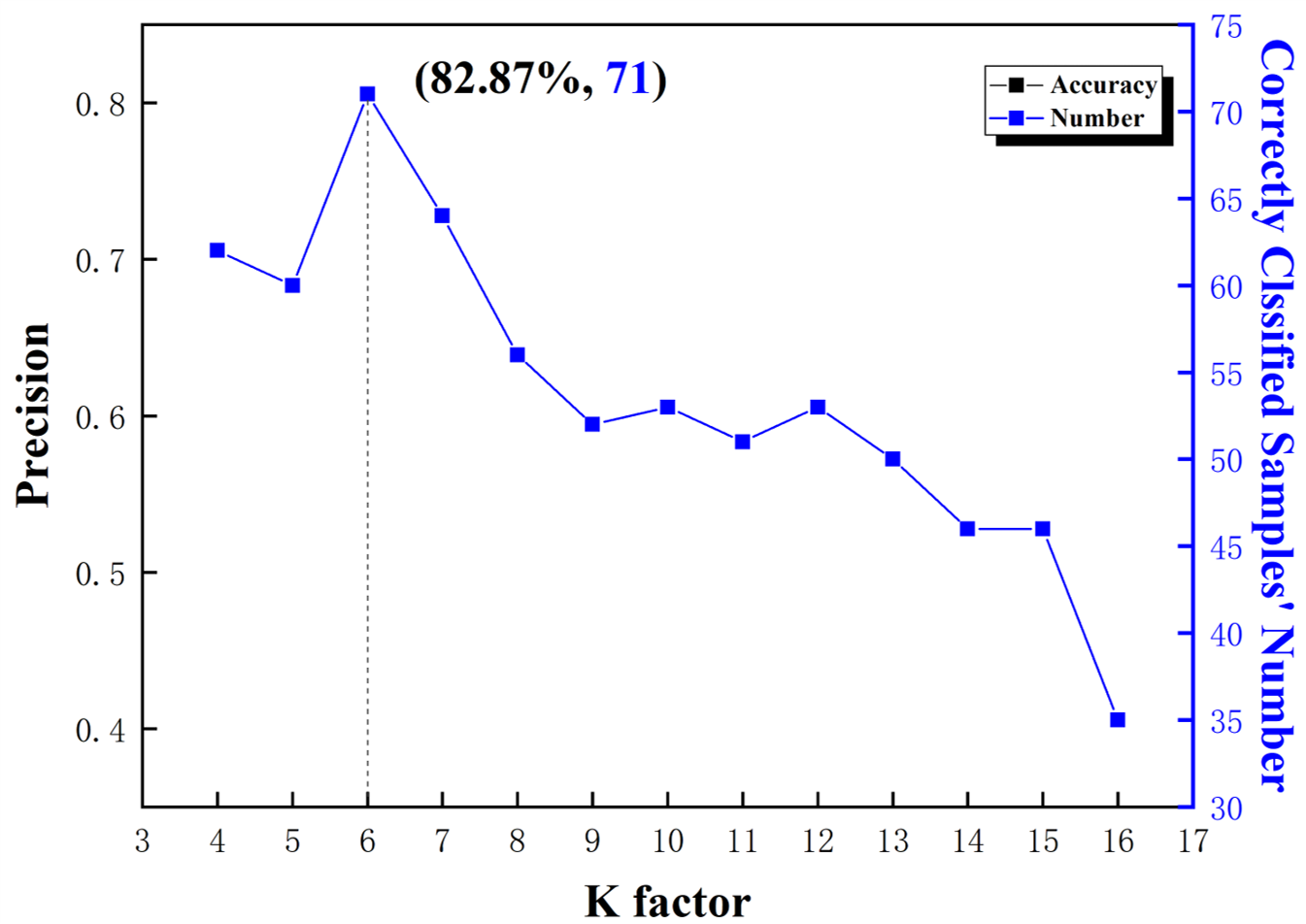}
	\caption{The influence of K factor on the performance of system. The Y axis on the right stands for the number of correctly predicted samples, and the Y axis on the left shows the corresponding accuracy. Intuitively, it can be seen that the optimal K is 6.}\label{F06}
\end{figure}
As mentioned above, we specially aim at obtaining the optimal $k$ to get the most effective detection system. In KNN algorithm, K plays a key role in the performance of system. In other words, if the K factor is too small, there may exist the following problems: 1) the phenomenon of over fitting; 2) being easy to influenced by outliers; and 3) the system is too complex. While the K factor is  too large, there may appear the elaborated problems: 1) the phenomenon of under fitting; 2) the system is constrained by the sample equilibrium; and 3) the system is too simple. 

Hence, in this study, we analyze the correctly classified samples' number and accuracy under the condition that K's values vary from 4 to 16, as demonstrated in Fig. \ref{F06}. Accordingly, we can find the optimal-k-value is 6, with the the number TP 71 and the precision 82.87\%. Additionally, K going from 4 to 6, the curve is roughly go up. For K ranging from 6 to 16, there are two stages in the descent tendency of the curve. First, the curve declines rapidly when K is from 6 to 9. Then it oscillates downward from 9 to 15, and the performance of system descends sharply when the K is changed to 16. The phenomenon referred above conforms to the theoretic prediction. Surprisingly, it is exactly consistent with the conclusion that the fixed optimal-k-value for all test samples should be $k=\sqrt{n}$ (where $n>100$ and $n$ is the number of training samples proposed by Lall and Sharma\cite{95WR02966}. 
\subsection{Comparison of measures in three types categories}

\begin{figure*}[ht]
	\centering
	\includegraphics[width=1\textwidth]{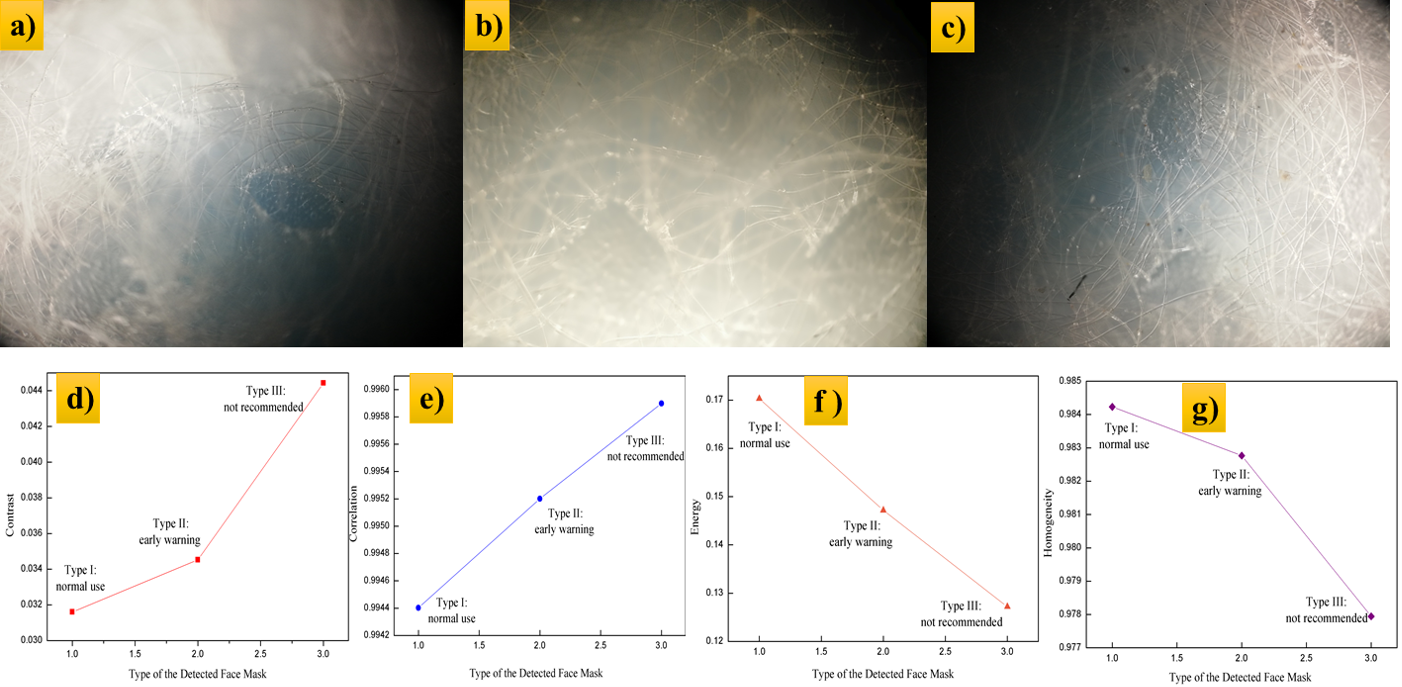}
	\caption{The micro-photos of three types face masks and their corresponding GLCMs' measures. a), b) and c) represent the face masks in `normal use' stage, `early warning' stage and `not recommended' stage, respectively. d), e), f) and g) are curve of four measures viz. contrast, correlation, energy and homogeneity.} \label{F07}
\end{figure*}
we analyze the measures of all samples to inquire the tendency of modeling measures over time changes. Fig. \ref{F07} displays the micro-photos and their corresponding GLCMs' measures of three different service stages viz. `normal use' stage, `early warning' stage and `not recommended' stage. As we can see from the micro-photos, when the use time increases, the mask becomes more and more dirty and there are many impurities and small droplets in the holes of the mask. It proves the accuracy of the analysis mentioned above that the protection becomes less effective over time. The curves reveal the variation tendency of the measures. Specifically, contrast and correlation increase while energy and homogeneity decrease over time. It should be noted that this rule can't be applied to all data, since the detection result is determined by composite force of the four measures. As a result, there exists the situation that some data don't obey the rule.

\subsection{Comparison of Experimental Results using original photos and micro-photos, respectively}
\begin{figure}[ht]
	\centering
	\includegraphics[width=0.45\textwidth]{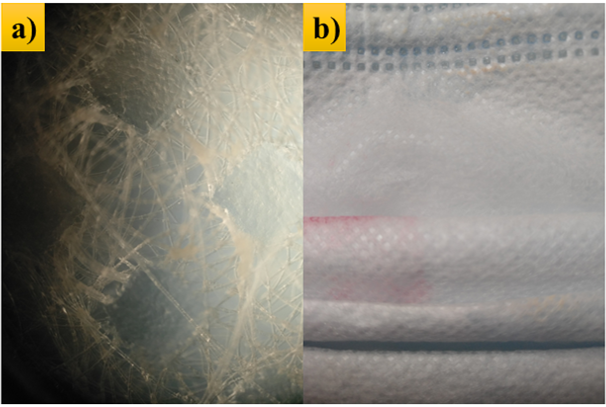}
	\caption{The original photo and the micro-photo of the same face mask. a) displays the micro-photo obtained by microscope while b) shows the original photo gained only by mobile phone's camera.}\label{F08}
\end{figure}

To validate the necessity of microscope, we record the results of using the original photo and micro-photo, which are shown in Fig. \ref{F08}. The two photos represent the same location of the face mask and belong to the type `not recommended'. In Fig. \ref{F08}, we can see many details, such as droplets in the holes, etc. Meanwhile, we can see no details in the original photo except the obvious stain. It can be inferred that we will tell little difference if there is no stain or other obvious dirt. The detection result of the micro-photo is `not recommended' while the original photo 's result is  `normal use', which verifies the indispensability of the microscope in this system. In other words, the photos obtained without using microscope aren't able to be used as the detection input.
\subsection{Dichotomous Model of Experimental Results} 

On the basis of system's working normally, we further have the idea whether we can simplify the system. As a result, we propose a dichotomous model with only two types viz. `normal use' and `not recommended'. Considering that `normal use' and  `not recommended' type face masks both can protect people effectively, we merge them into one type named `normal use'. Accordingly, the other type is `not recommended'. In the test samples, the number of `normal use' is 60 (54 samples are detected correctly) and `not recommended' is 27 (23 samples are detected correctly). To counterbalance the uneven distribution of samples, we assign the weight viz. $\frac{60}{87}$ and $\frac{27}{87}$ to each type. As expected, the accuracy becomes better and increases to 88.51\%. Since there are two types, it is easier for the model to tell which type the detected face mask belongs to. In some simple cases, we may consider this scheme.
\subsection{Limitation of face mask condition }
Besides the system mentioned above being able to be used to detect its service life, some conditions of the face mask itself may also affect their use life. For example, there may be lint inside the face mask when using face mask a few times. In that case, we can choose to discard the mask and use a new one to ease the discomfort. Meanwhile, if the face mask being used gets wet by accident, we must stop using it whatever stage it belongs to.

\section{CONCLUSION}
\label{sec3}

In this paper, we propose a service stage detection method based on a mobile microscope which can be used to obtain the micro-photos of the face mask being used.  

In our detection method, we first get the micro-photos of face mask being used, which may reflect some details such as droplets or other obvious dirt. Then, we extract texture features from the micro-photos using the method of GLCM and choosing four measures viz. contrast, correlation, energy and homogeneity as the features. Subsequently, KNN method is applied to work on the detect the service stage. In validation experiments, the obtained system achieves a relatively good result with a precision of 82.87\%$\pm$8.50\% on the testing dataset. The result of our experiments provide an idea of using photos to detect the use time of face masks to distinguish whether the face mask can be used sequentially or not. Faced with the current severe situation of COVID-19 and possible shortage of face masks, our research may work as an assistant to help the common people to use face masks more correctly. Furthermore, it can exert positive influence on protecting uninfected people and stopping the possibly infected patients from spreading virus.

In future research, based on the current portability and practicability, we will try using a more stable algorithm to reach the goal of achieving a higher accuracy. Besides, different types of face masks should be taken into consideration to expand the types of detecting objects.

\small
\bibliographystyle{IEEEbib}
\bibliography{strings,refs}

\end{document}